\documentclass[manuscript,screen]{acmart}

\usepackage{amsmath,multirow}
\usepackage{graphicx}
\usepackage{algorithm}
\usepackage{algorithmic}
\usepackage{subfigure,color}
\newcommand{\hao}[1]{\textcolor{black}{#1}}
\newcommand{\tht}[2]{\begin{tabular}{@{}#1@{}}#2\end{tabular}}
\usepackage{gensymb,graphics,subfigure,inputenc}
\graphicspath{{Figures/}}
\usepackage{textcomp,subfigure,multirow,upgreek}
\usepackage{booktabs}
\usepackage{color,mdwlist}
\usepackage{lineno,hyperref}

\AtBeginDocument{%
  \providecommand\BibTeX{{%
    \normalfont B\kern-0.5em{\scshape i\kern-0.25em b}\kern-0.8em\TeX}}}

\setcopyright{acmcopyright}
\copyrightyear{2018}
\acmYear{2018}
\acmDOI{10.1145/1122445.1122456}

\acmConference[Woodstock '18]{Woodstock '18: ACM Symposium on Neural
  Gaze Detection}{June 03--05, 2018}{Woodstock, NY}
\acmBooktitle{Woodstock '18: ACM Symposium on Neural Gaze Detection,
  June 03--05, 2018, Woodstock, NY}
\acmPrice{15.00}
\acmISBN{978-1-4503-XXXX-X/18/06}

\begin{document}

\title{Deep Unsupervised Key Frame Extraction for Efficient Video Classification}

\author{Hao Tang}
\email{hao.tang@vision.ee.ethz.ch}
\authornotemark[1]
\affiliation{%
  \institution{ETH Zurich}
\country{Switzerland}
}

\author{Lei Ding}
\affiliation{%
  \institution{University of Trento}
  \country{Italy}}

\author{Songsong Wu}
\affiliation{%
	\institution{Guangdong University of Petrochemical Technology}
	\country{China}}

\author{Bin Ren}
\affiliation{%
	\institution{University of Trento}
	\country{Italy}}

\author{Nicu Sebe}
\affiliation{%
	\institution{University of Trento}
	\country{Italy}}

\author{Paolo Rota}
\affiliation{%
	\institution{University of Trento}
	\country{Italy}}

\renewcommand{\shortauthors}{Hao Tang, Lei Ding, Songsong Wu, Bin Ren, Nicu Sebe, Paolo Rota}

\begin{abstract}
Video processing and analysis have become an urgent task since a huge amount of videos (e.g., Youtube, Hulu) are uploaded online  every day. The extraction of representative key frames from videos is very important in video processing and analysis since it greatly reduces computing resources and time. Although great progress has been made recently, large-scale video classification remains an open problem, as the existing methods have not well balanced the performance and efficiency simultaneously. To tackle this problem, this work presents an unsupervised method to retrieve the key frames, which combines Convolutional Neural Network (CNN) and Temporal Segment Density Peaks Clustering (TSDPC). The proposed TSDPC is a generic and powerful framework and it has two advantages compared with previous works, one is that it can calculate the number of key frames automatically.  The other is that it can preserve the temporal information of the video.  Thus it improves the efficiency of video classification. Furthermore, a Long Short-Term Memory network (LSTM) is added on the top of the CNN to further elevate the performance of classification. Moreover, a weight fusion strategy of different input networks is presented to boost the performance. By optimizing both video classification and key frame extraction simultaneously, we achieve better classification performance and higher efficiency. We evaluate our method on two popular datasets (i.e., HMDB51 and UCF101) and the experimental results consistently demonstrate that our strategy achieves competitive performance and efficiency compared with the state-of-the-art approaches.

\end{abstract}




\maketitle

\section{Introduction}

\begin{figure*}
	\centering
		\includegraphics[width=1\linewidth]{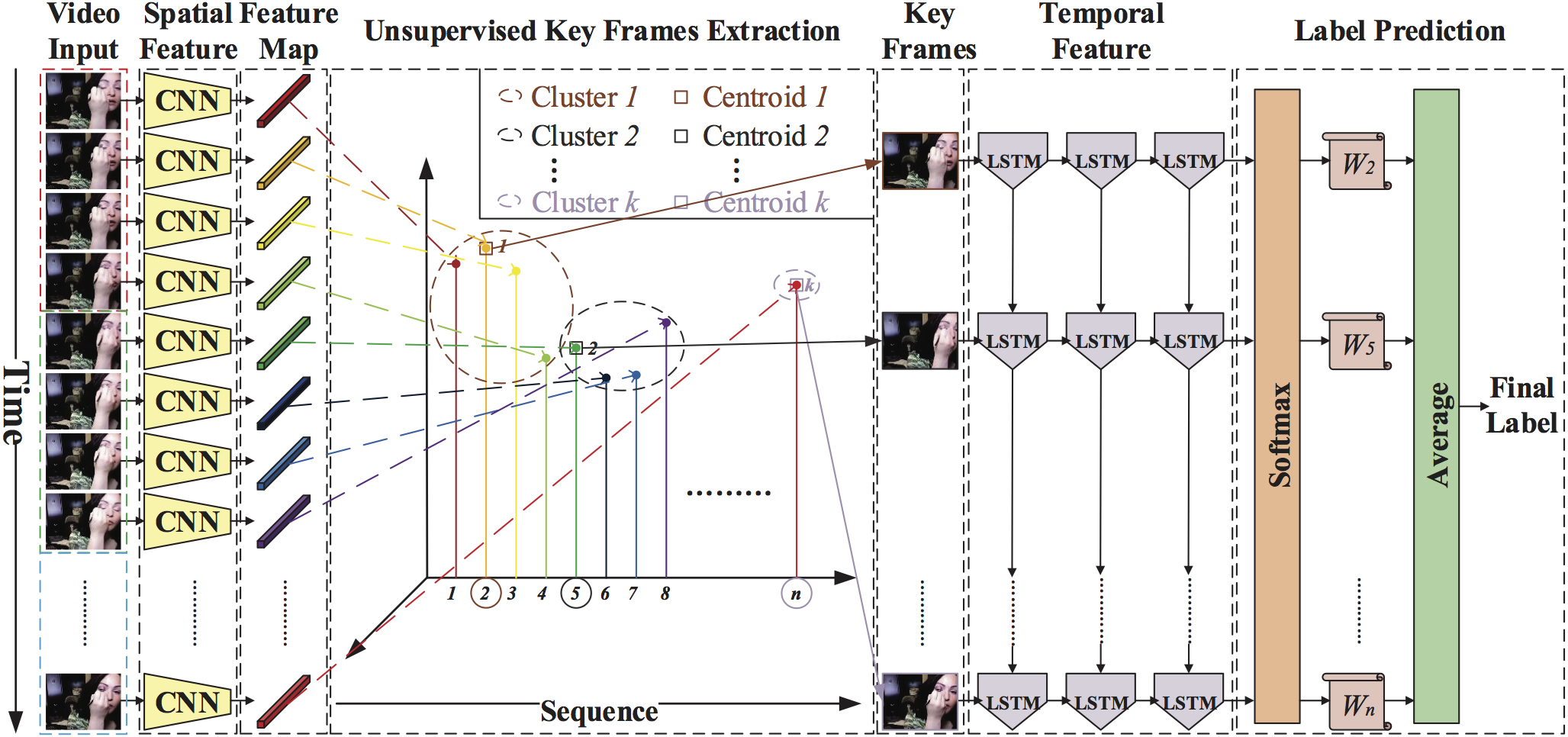}
		\caption{The pipeline of the proposed large-scale video classification, which has three components, i.e., CNN, TSDPC and LSTM. The TSDPC is an unsupervised key frame extraction method which preserves the temporal information of the video and calculates the the number of key frames automatically. The CNN-TSDPC is proposed for accelerating video processing. In addition, a LSTM cell is further connected at the end of the CNN, which is able to elevate the performance of video classification. 
		At last, the LSTM predicts the video label at each key frame and we average these predictions for final labels.}
		\label{fig:framework}
\end{figure*}

Recent years have witnessed an exponential growth in video data availability on the web, such as Youtube. 
In this situation, large-scale video classification techniques \cite{feichtenhofer2018have,feichtenhofer2017spatiotemporal,carreira2017quo,feichtenhofer2016spatiotemporal,wang2017spatiotemporal,dutaspatio,roberto2017procedural,wang2016temporal,kar2016adascan,zhu2016key,hara2018can,zhou2018mict,gao2017im2flow,wang2018temporal,zhu2018towards,tang2018fast,choutas2018potion,tran2018closer,wu2018compressed,liu2015sdm,liu2016sequential}
 have also received increasing interests due to the fact that there is an increasing demand for efficient indexing and managing of these video data.

Efficient and accurate large-scale video classification performance relies on the extraction of discriminative spatial and temporal features. 
Conventional approaches extract information or feature using hand-crafted features (e.g., Histogram of Optical Flow (HOF)~\cite{Laptev2008Learning} or improved Dense Trajectories (iDT)~\cite{wang2013action}) from video frames which are then encoded (e.g., Fisher Vector (FV)~\cite{wang2013action}) and pooled (e.g., average pooling) to produce a global feature representation and then passed to a classifier (e.g.,~SVM). 

In recent years, video classification research has been influenced by the trends of deep learning.
A common pipeline of these works is to use a group of frames as the input to the network, whereby the model is expected to learn spatio-temporal features. 
Simonyan et al.~\cite{simonyan2014two} propose a two-stream ConvNet architecture which incorporates spatial and temporal networks to extract spatial and temporal features.
Feichtenhofer et al.~\cite{feichtenhofer2016convolutional} study a number of ways of fusing ConvNet towers both spatially and temporally in order to best take advantage of this spatio-temporal information.
Wang et al.~\cite{wang2016temporal} present the Temporal Segment Network (TSN), which combines a sparse temporal sampling strategy and video-level supervision to enable efficient and effective learning using the whole action video.
Ji et al.~\cite{ji20133d} introduce a 3D CNN model, in which both spatial and temporal features are extracted by performing 3D convolution. 
Tran et al.~\cite{tran2015learning} propose an approach for spatio-temporal feature learning using deep 3-dimensional convolutional networks (3D ConvNets).

However, these approaches for video classification treat the videos in a holistic way, i.e., as one data instance. 
Feature representation learned from the entire video unavoidably brings redundant information from the repeated and unrelated video frames which leads to two problems,
(i) one is that processing the whole video data requires excessive computational resources and time, for example, Kulhare et al.~\cite{kulhare2016key} report that the optical flow data on 13K videos was 1.5 TB.
From the computational time perspective, these methods require extensively long periods of training time to effectively optimize millions of parameters that represent the model;
(ii) another problem is that repeated and unrelated video frames such as blurry, feature-less and background frames always overwhelm the targeted visual patterns, which sometimes confuse the classifier. 

To fix these problems, many efforts have been done to explore the key frame extraction methods which can convert video processing to image processing.
For example, Donahue et al.~\cite{donahue2015long} employ a Long Short-Term Memory network (LSTM) which is connected to the output of the underlying CNN.
However, this framework processes 16 sample frames selected evenly with a stride of 8 frames from the full length video as the video representation. 
Such average selection of samples may not consider all useful motion and spatial information. 
Besides, Zhu et al.~\cite{zhu2016key} propose a key volume mining deep framework to identify key volumes to achieve high accuracy, but with a steep increase of the computational load.

To solve both limitations, we present a novel framework to model video representation in an efficient and accurate manner.
Instead of trying to learn features over the entire video as in previous works, which always contain redundant information, resulting in degraded performance. 
We consider a different manner to extract video features over several key frames of the entire video, which is a fundamental way to alleviate the computation burden.
Key frames, also named representative frames, are often employed to represent the story of a video.
Key frame extraction is the key technology for video abstraction, which can remove the redundant information in the video and reduce the high computing redundancy between adjacent frames.
The algorithm for key frame extraction will affect the reconstruction of video content.

To this end, we propose an unsupervised key frame extraction method by using Convolutional Neutral Networks (CNNs) and Density Peaks Clustering (DPC)~\cite{rodriguez2014clustering}. 
More specifically, we extract convolutional deep feature for each frame in the video and then map these deep feature maps into a high dimensional feature space.
After that, how to describe the feature space is a hard nut to crack for its uneven distribution.
To solve this, we propose the Temporal Segment Density Peaks Clustering (TSDPC) to select key frames in an unsupervised way in feature space.
In order to preserve the temporal cues of the video, we first segment it into several segments, then select key frames on each segment using DPC.
Finally, we combine  all the key frames extracted from each segment to form the final key frames.
TSDPC has two advantages compared with previous clustering-based method,
(i) Previous clustering approaches cannot detect non-spherical clusters due to the fact that they only rely on the distance between feature points to do clustering.
While TSDPC is the approach based on the local density of feature points, which is able to detect non-spherical clusters. 
(ii) Clumsy tricks are used to ensure the number of key frames which may bring uncertainty in previous works, while for TSDPC, it can calculate the number of key frames automatically.

After extracting key frames, we replace the original video sequence with the key frames as a surrogate for analyzing, which could greatly enhance the time efficiency with little cost of accuracy.
In addition, in order to improve the accuracy, a Long Short-Term Network (LSTM) is further connected at the end of the CNN.
Furthermore, a novel input network fusion strategy with different weights is presented.
The pipeline of the proposed large-scale video classification framework is shown in Figure~\ref{fig:framework}.
Finally, experimental results show that the proposed framework is accurate and efficient for video classification on two public datasets.

The contributions of this work can be summarized as follows:
\begin{itemize} 
	\item We propose a novel method of unsupervised key frames extraction in video, CNN-TSDPC, which is comprised of CNN and Temporal Segment Density Peaks Clustering (TSDPC). Note that the proposed TSDPC can preserve the temporal information of a video and calculate the number of key frames automatically.
	\item The proposed CNN-TSDPC-LSTM framework is trained in an end-to-end fashion to improve both performance and efficiency for the video classification task.
	\item We present a weight fusion strategy of different input networks, in which different inputs are fused with different weights to elevate the accuracy of video classification.
	\item The proposed method and framework achieve competitive performance and are more efficient compared with the state-of-the-art models.
\end{itemize}

\section{Related Work}

Video content summarization has been widely used to facilitate the indexing of large videos. 
In earlier works on video summarization, key frames are selected either by sampling video frames randomly or uniformly at certain time intervals.
To make a key frame extraction algorithm effective, the extracted key frames should represent the whole video content without missing important information such as people and objects.
While at the same time, video content information of these key frames should not be similar, tin order to avoid content redundancy.
In recent years, many algorithms have been proposed for key frame extraction in videos . 
There are three categories of key frames extraction algorithms.

\noindent \textbf{Segmentation Based.}
These methods detect abrupt changes in terms of similarity between successive frames. 
The key frames are equidistant in the video curve with respect to Iso-Content Distance, Iso-Content Error and Iso-Content Distortion in~\cite{panagiotakis2009equivalent}.
In~\cite{ejaz2012adaptive}, a key frame is extracted if and only if the inter-frame difference overtakes a certain threshold. 
A key frame selection method based on key points is presented in~\cite{guan2013keypoint}, in which
a global pool of key points based on SIFT feature extracted from all frames is generated and those frames that best cover the global key point pool are selected as key frames. 
In a word, this type of key frame extraction methods have the disadvantage, i.e., it may extract similar key frames if the same content reappears during a video.

\noindent \textbf{Dictionary Based.}
These types of approaches convert key frames extraction into a sparse dictionary selection problem.  
In~\cite{cong2012towards}, a key frame selection method is proposed which is based on the sparse dictionary selection with the loss in $L_{2,1}$ norm. 
Besides, the key frames are selected based on the true sparse constraint $L_0$ norm to represent the whole video in~\cite{mei2015video}.
$L_{2,0}$ constrained sparse dictionary selection model is proposed to solve the problems that the  solution based on the convex relaxation  cannot guarantee the sparsity of the dictionary and it selects key frames in a local point of view in~\cite{mei20142}.
In~\cite{meng2016keyframes}, a video is summarized into a few key objects by selecting representative object proposals generated from video frames based on the sparse dictionary selection method.
To find key frames with both diversity and representativeness, the objective function in~\cite{wang2017representative} consists of a reconstruction error and three structured
regularizers, i.e., group sparsity regularizer, diversity regularizer, and locality-sensitivity regularizer. 

\noindent \textbf{Clustering Based.}
These approaches cluster frames into groups and then select the frames closest to the cluster centers as key frame. 
In~\cite{zhuang1998adaptive}, key frames are detected using unsupervised clustering based on visual variations. 
While in~\cite{kuanar2013video}, dynamic delaunay clustering is adopted to extract key frames.
Key frames are selected based on color feature using the k-means clustering algorithm in~\cite{de2011vsumm}. 
In~\cite{vazquez2013spatio}, spectral clustering on spatio-temporal features is employed to extract key frames. 
Moreover, the mutual information values of these consecutive frames are clustered into several groups using a split-merge method in~\cite{cernekova2006information}. 
In~\cite{panda2014scalable}, the problem is modeled as a graph clustering problem and it is solved using a skeleton graph.
Besides, the authors present a key frame extraction approach based on local description and graph modularity clustering in~\cite{gharbi2017key}.

However, the proposed CNN-TSDPC framework has three advantages compared with these approaches,
(i) our framework can capture the discriminative spatial information by using a sophisticated CNN feature extractor.
(ii) our framework can represent the temporal information by using a temporal segmentation strategy.
(iii) our framework can extract more representative key frames with less redundant information than previous works.

\section{Method}

\hao{We assume that a video $\textbf{\text{V}}$ can be divided into $K$ segments $\{V_1, V_2, \cdots, V_K\}$ of equal durations.}
For each segment $V_K$, we assume that a frame in $V_K$ can be represented by
$\textbf{\text{f}}_i^K$, where $i \in [1, 2, \cdots, N]$ and $N$ is the number of frames in $V_K$,
\begin{equation}
{V_K} = \{\textbf{\text{f}}_i^K\}_{i=1}^N.
\end{equation}
Hence, the key frame set $\textbf{\text{f}}_{m_k}^K$ is defined as follows:
\begin{equation}
\label{equ:keyfra}
\textbf{\text{f}}_{m_k}^K = \varTheta({V_K}), k \in [1, 2, \cdots, n_c],
\end{equation}
where $m_k $ is the index of the key frames, $n_c \geq 1$ is the number of key frames and $\varTheta$ denotes the key frames extraction procedure, the objective is to remove the redundant data which will significantly reduce the amount of information to be processed. 
It is necessary to discard the frames with repetitive or redundant information during the extraction. 
Thus, key frame extraction is the fundamental step in video analysis applications.

\subsection{Convolutional Deep Feature Extraction}
We try to find a proper descriptive index to evaluate each frame in a segment ${V_K}$, facilitating key frame extraction.
Informative frames could better summarize the whole video, but how to quantify the information each frame contains is a hard nut to crack.
Previous works such as~\cite{de2011vsumm} adopt the color feature to represent each frame.
\cite{tang2018fast} uses the image entropy as a feature representation for hand action recognition.
However, we argue that these feature extractors are not powerful to extract discriminative information.
In this paper, we use CNNs to extract the deep feature.
CNNs are powerful due to their ability to extract the semantic features of an image. 
The first few convolutional layers can identify lines and corners, and then we pass these patterns down through more convolutional layers and start recognizing more complex and abstract semantic features as we go deeper. 
This property makes CNNs really good at extracting features in images and videos.

For each frame $\textbf{\text{f}}_i^K$ in the segment ${V_K}$, we adopt ResNet~\cite{he2016deep} as deep feature extractor. 
This pre-trained model of ResNet is trained on a subset of the ImageNet dataset~\cite{russakovsky2015imagenet}. 
The model is trained on more than one million images and can classify images into 1,000 object categories. 
The feature vector $\textbf{\text{x}}_i^K$ can be obtained as follows:
\begin{equation}
\label{equ:extract_cnn}
	\textbf{\text{x}}_i^K = \varDelta(\textbf{\text{f}}_i^K), i\in [1, 2, \cdots , N]
\end{equation}
where $\varDelta$ denotes CNN feature extraction operation and $\textbf{\text{x}}_i^K$ is the corresponding feature vector of frame $\textbf{\text{f}}_i^K$.
After extracting deep features of each frame via feature extractor, feature vectors $\textbf{\text{x}}_i^K, i\in [1, 2, \cdots , N]$ are then mapped to the points in high dimension feature space $\mathcal{F}$. 

\subsection{Temporal Segment Density Peaks Clustering}
We define the following symbols for the sake of simplicity,
\begin{equation}
S = \{\textbf{\text{x}}_i^K\}_{i=1}^N, I_S = \{1, 2, . . . , N\}.
\end{equation}
We consider that the distribution of $S$ in $\mathcal{F}$ should have the following two characteristics: (i) the cluster centers of $S$ are surrounded by neighbors with a lower local density; (ii) these centers have a relatively large distance from any points with a higher local density.
Thus, we adopt density peaks clustering to further cluster these feature vector $\textbf{\text{x}}_i^K$.
Density peaks clustering~\cite{rodriguez2014clustering} could better catch the delicate spherical structure of space where points reside than traditional clustering strategies, e.g., K-means, in which features are grouped to the nearest cluster center.

For each point $\textbf{\text{x}}_i^K$ in $S$, we compute two quantities: the local density $\rho_i$ and its distance $\delta_i$ from points of higher density.
Both of these quantities depend only on the distance $d_{ij}$ between points in $\mathcal{F}$, 
\begin{equation}
d_{ij} = dist(\textbf{\text{x}}_i^K, \textbf{\text{x}}_j^K)
\end{equation}
The local density $\rho_i$ of point $\textbf{\text{x}}_i$ is defined as:
\begin{equation}
\label{equ:cutoff}
\rho_i = \sum_{j \in \{{I_S - \{i\}}\}} \chi(d_{ij} - d_c),
\end{equation}
where,
\begin{equation}
\chi(x) = 
\begin{cases}
1, & x <0;\\
0, & x \geq 0,
\end{cases}
\end{equation}
and $d_c$ is a cutoff distance.
Basically, $\rho_i$ equals to the number of points that are closer than $d_c$ to point $\textbf{\text{x}}_i^K$.
The algorithm is sensitive only to the relative magnitude of $\rho_i$ in different points, which implies that, the results of the analysis are robust with respect to the choice of $d_c$ for large dataset.
A different way to define $\rho_i$ is:
\begin{equation}
\label{equ:gaussian}
\rho_i = \sum_{j \in \{{I_S - \{i\}}\}} e^{-(\frac{d_{ij}}{d_c})^2},
\end{equation}
in which a Gaussian kernel is used to calculate the local density.
We can see from these two definitions that the cutoff kernel in Eq.~\eqref{equ:cutoff} is a discrete value, while Gaussian kernel in Eq.~\eqref{equ:gaussian} is a continuous value, which guarantees a smaller probability of conflict.
In other words, the probability that different point have the same local densities, $\rho_i$, is smaller.

Another important quantity is $\delta_i$, which is measured by the minimum distance between the point $\textbf{\text{x}}_i^K$ and any other point with higher density:
\begin{equation}
\label{equ:delta}
\delta_i = 
\begin{cases}
\min\limits_{j \in I_S^i}\{d_{ij}\}, & I_S^i \neq \varnothing;\\
\max\limits_{j \in I_S}\{d_{ij}\},   & I_S^i = \varnothing,
\end{cases}
\end{equation}
where, 
\begin{equation}
I_S^i = \{k \in I_S : \rho_k > \rho_i\}.
\end{equation}
Obviously, the $I_S^i = \varnothing$ if $\rho_i = \max\limits_{j \in I_S}{\rho_j}$.

\begin{figure*}
	\centering
	\setcounter{subfigure}{0}
	\subfigure[The HMDB51 dataset.]        {\includegraphics[width=0.4\linewidth]{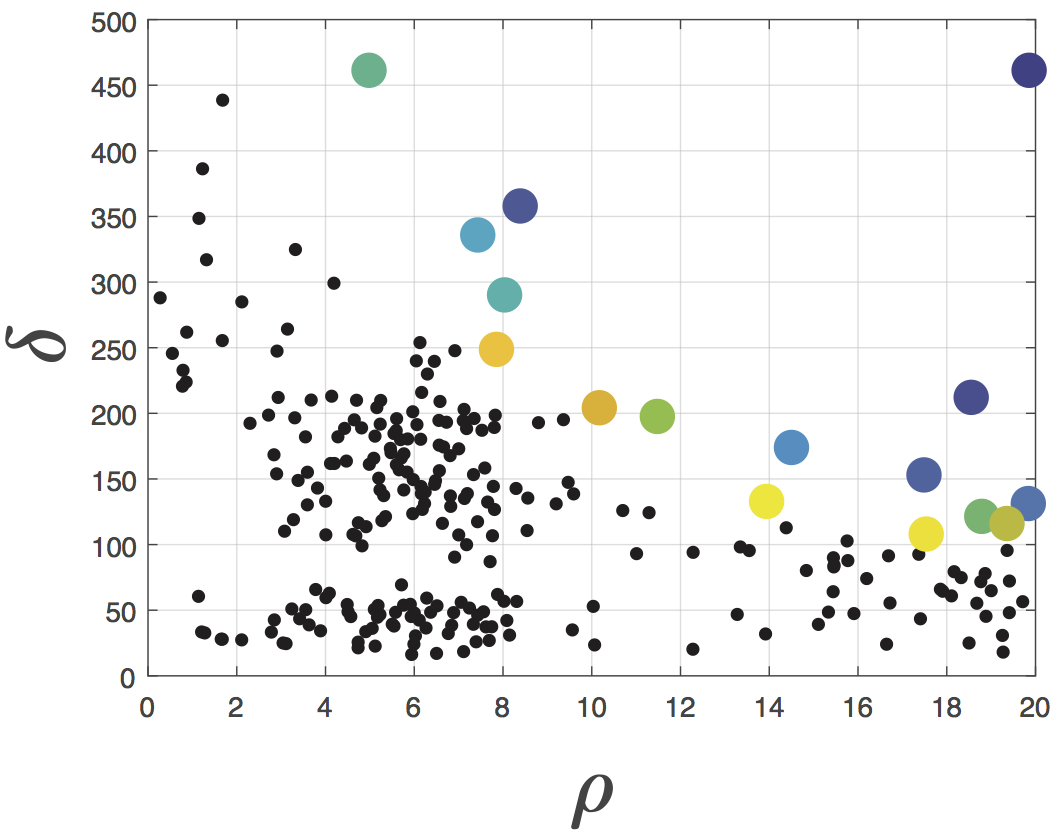}}
	\subfigure[The HMDB51 dataset.]        {\includegraphics[width=0.4\linewidth]{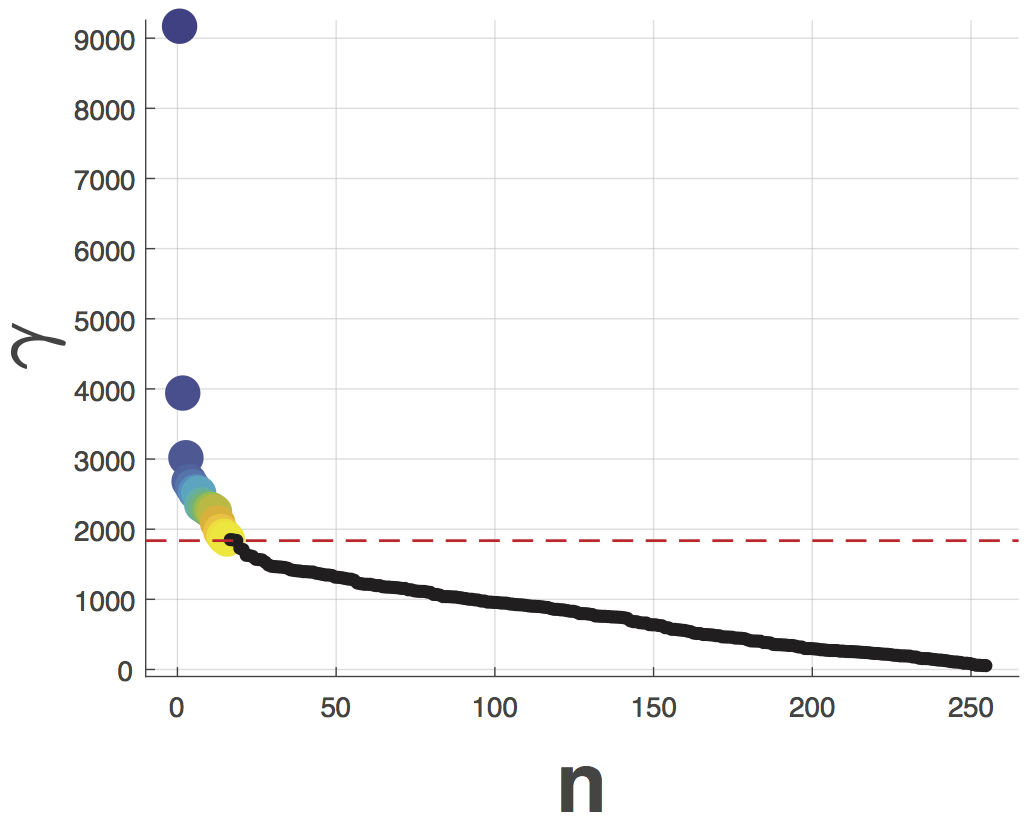}} 
	\subfigure[The UCF101 dataset.]        {\includegraphics[width=0.4\linewidth]{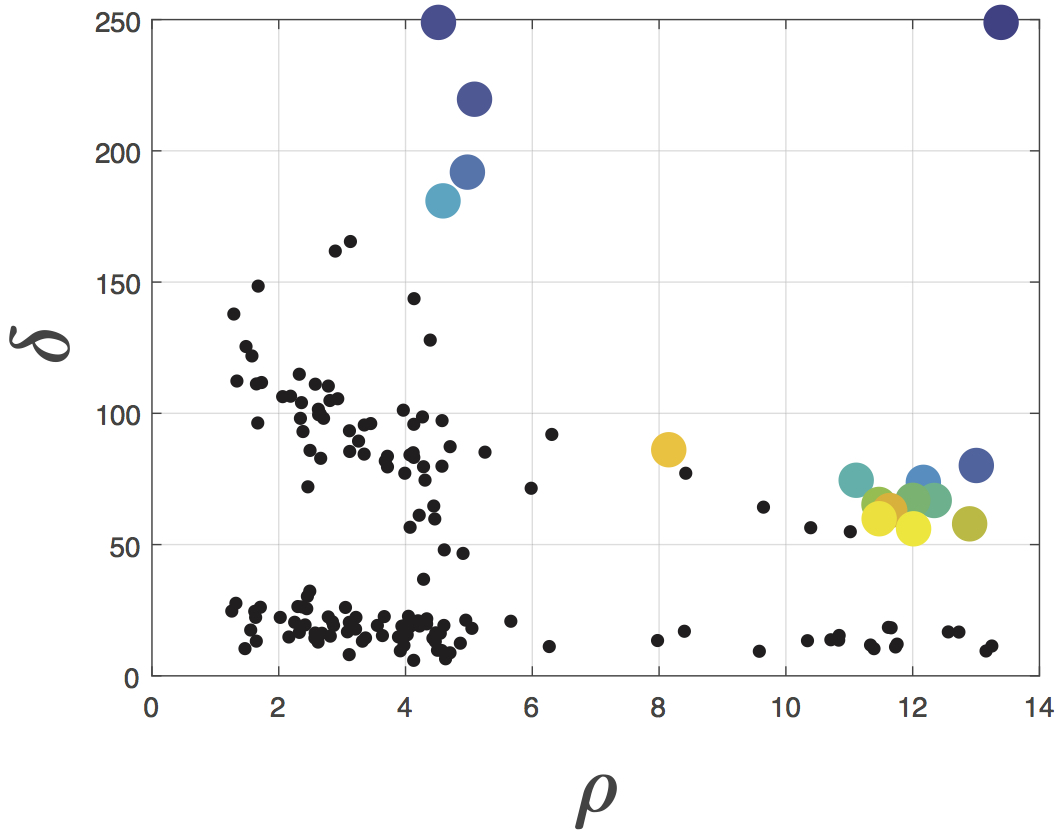}}
	\subfigure[The UCF101 dataset.]        {\includegraphics[width=0.4\linewidth]{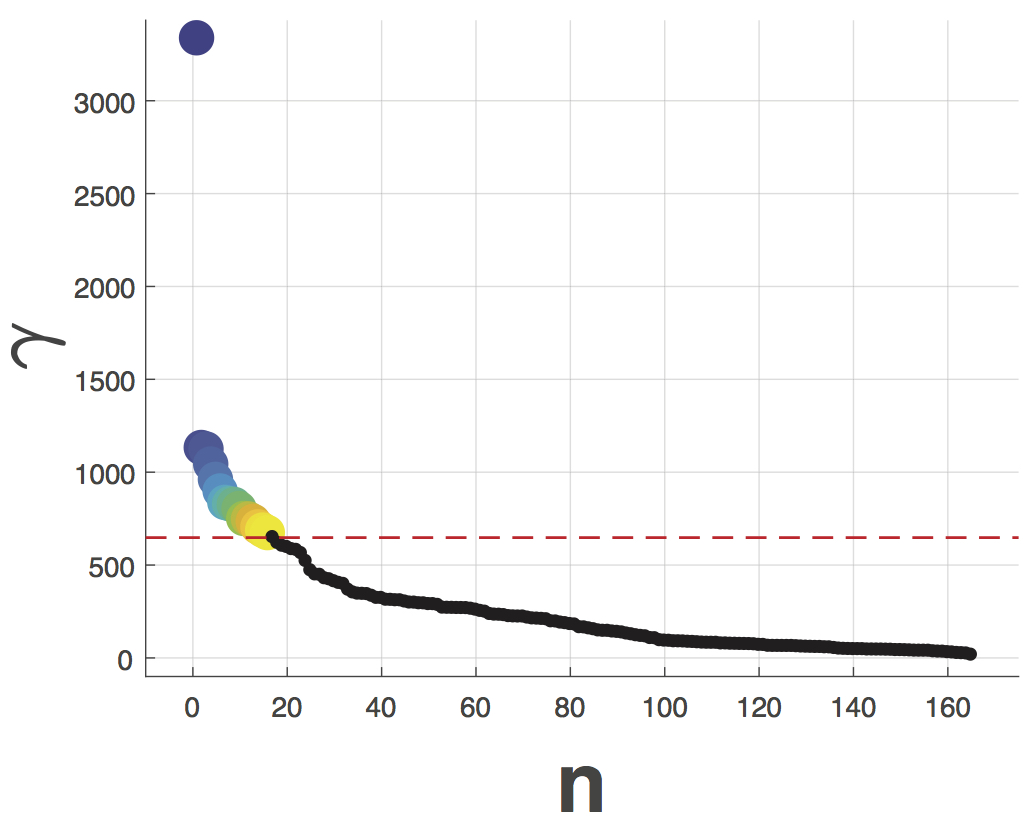}} \\
	\caption{$\rho$, $\delta$ and $\gamma_i$ on HMDB51 and UCF101 datasets, respectively. 
	}
	\label{fig:papameter}
\end{figure*}

Consequently, for each point $\textbf{\text{x}}_i^K$ in the $S$, we can calculate binary pair $(\rho_i, \delta_i)$, where $i \in I_S$.
The definition of quantity $\gamma_i$ which considers both $\rho_i$ and $\delta_i$ is as follows,
\begin{equation}
\label{equ:gamma}
\gamma_i = \rho_i \delta_i, i \in I_S,
\end{equation}
we select the point with the larger value of $\gamma_i$ as the cluster center.
Figure~\ref{fig:papameter} shows the selection of $\rho$, $\delta$ and $\gamma_i$ on two videos of HMDB51 and UCF101 datasets, respectively.

In experiments, according to previous works on temporal modeling~\cite{wang2016temporal,gaidon2013temporal,wang2014latent} we set $K$ to 3.
Since we observe that when $K=1$, the key frames extracted by the proposed method cannot preserve the temporal information in the whole input video.
We do the same operations on other segments and combine all the key frames from each segment to form the final key frames~$\textbf{\text{f}}=\{\textbf{\text{f}}^1, \textbf{\text{f}}^2, \cdots, \textbf{\text{f}}^K\}$ of the video $\textbf{\text{V}}$.

The pipeline of the proposed CNN-TSDPC framework is summarized in Algorithm~\ref{alg:cdpc}.
The CNN-TSDPC method is comprised of three steps.
The first step is to segment a video into several volumes equally (step 1.1).
The second step is convolutional deep feature extraction, we extract deep feature vector $\textbf{\text{x}}_i$ using Eq.~\eqref{equ:extract_cnn} and then map feature vector $\textbf{\text{x}}_i$ to feature space $\mathcal{F}$ (step 2.1 and step 2.2).
After that, we conduct density peaks clustering operation.
We first calculate the distance $d_{ij}$ between $\textbf{\text{x}}_i$ and $\textbf{\text{x}}_j$, and then calculate the cut off distance $d_c$ with the given $t$ (step 3.1 and step 3.2).
\hao{According to the asseverations in~\cite{rodriguez2014clustering}, $d_c$ can be selected as the rule that  the average number of neighbors is around 1\% to 2\% of the total number of points in the data set. }
In order to obtain $\gamma_i$, we need to calculate $\rho$ and $\delta$ with Eq.~\eqref{equ:cutoff} or Eq.~\eqref{equ:gaussian} and Eq. \eqref{equ:delta} (step 3.3 and step 3.4).
And then multiply $\rho$ and $\delta$ to obtain $\gamma$ using Eq.~\eqref{equ:gamma} (step 3.5).
Next, we rank $\gamma_i$ in descending order and choose the $n_c$ largest $\gamma$ values as the clustering centers.
In the end of our algorithm, the index of the key frames $m_k$ and key frames $\{\textbf{\text{f}}_{m_k}\}_{k=1}^{n_c}$ of ${V_K}$ are returned for further processing.

\begin{algorithm}
	\caption{The pipeline of the proposed CNN-TSDPC method.}
	\label{alg:cdpc}
	\begin{algorithmic}
		\REQUIRE The video $\textbf{\text{V}}$.
		\ENSURE The key frames $\textbf{\text{f}}=\{\textbf{\text{f}}^1, \textbf{\text{f}}^2, \cdots, \textbf{\text{f}}^K\}$.
		\STATE \textbf{Step 1} Temporal segmentation. \\
		\quad \textbf{1.1} Divide a video $\textbf{\text{V}}$ into $\{V_1, V_2, \cdots, V_K\}$, for each segment in $\{V_1, V_2, \cdots, V_K\}$, we do the following steps. \\		
		\STATE \textbf{Step 2} Convolutional deep feature extraction. \\
		\quad \textbf{2.1} Extract deep feature vector $\textbf{\text{x}}_i^K \leftarrow$ Eq.~\eqref{equ:extract_cnn}.  \\
		\quad \textbf{2.2} Map $\textbf{\text{x}}_i^K$ to feature space $\mathcal{F}$. \\			
		\STATE \textbf{Step 3} Density peaks clustering.\\
		\quad \textbf{3.1} Calculate $d_{ij}$ and $d_{ij} = d_{ji}, i<j, i,j \in I_S$.\\
		\quad \textbf{3.2} Given parameter $t \in (0, 1]$ to calculate $d_c$,
		\begin{equation}
			d_c = d_{f(Mt)},
		\end{equation}
		where $f(Mt)$ denotes the integer after rounding off $Mt$ and $M = \frac{1}{2}N(N-1)$ and $d_1 \leq d_2 \leq \cdots \leq d_M$. \\
		\quad \textbf{3.3} Calculate $\rho_i \leftarrow$ Eq.~\eqref{equ:cutoff} or Eq.~\eqref{equ:gaussian}.\\
		\quad \textbf{3.4} Calculate $\delta_i \leftarrow$ Eq.~\eqref{equ:delta}.\\
		\quad \textbf{3.5} Calculate $\gamma_i \leftarrow$ Eq.~\ref{equ:gamma}.\\
		\RETURN Index of the key frames $m_k$ and key frames $\{\textbf{\text{f}}_{m_k}^K\}_{k=1}^{n_c}$.
	\end{algorithmic}
\end{algorithm}

\section{Large-Scale Video Classification}
\label{sec:featurefusionstrategy}
 
To perform the sequential image classification, we employ Long Short-Term Memory networks (LSTMs), which is added at the top of CNNs.
LSTMs have been widely used to advance the state-of-the-art of many difficult problems because they are effective at capturing long term temporal dependencies.
We adopt the LSTM unit in our large-scale video classification framework, which is comprised of six important parts, i.e., block input, input gate, forget gate, cell state, output gate and block output.
Let $\odot$ denote the point-wise multiplication of two vectors, and let $\sigma(x) = \frac{1}{1 + e^{-x}}$ be the sigmoid non-linear activation function which squashes real-valued inputs to a $[0, 1]$ range, and let $\phi = \frac{e^x - e^{-x}}{e^x + e^{-x}} = 2\sigma(2x) - 1$ be the hyperbolic tangent nonlinearity activation function, which can also squash its inputs to the range of $[-1, 1]$. 
The LSTM in time step $t$ given inputs $x_t$ and $c_{t-1}$ is defined as follows:
\begin{equation}
g_t = \sigma(W_{xc}x_t + W_{hc}h_{t-1} + b_c),
\end{equation}

\begin{equation}
i_t = \sigma(W_{xi}x_t + W_{hi}h_{t-1} + b_i),
\end{equation}

\begin{equation}
f_t = \sigma(W_{xf}x_t + W_{hf}h_{t-1} + b_f),
\end{equation}

\begin{equation}
c_t = i_t \odot g_t + f_t \odot c_{t-1},
\end{equation}

\begin{equation}
o_t = \sigma(W_{xo}x_t + W_{ho}h_{t-1} + b_o),
\end{equation}

\begin{equation}
h_t = o_t \odot \phi(c_t).
\end{equation}
In this paper, three layers of LSTM is adopted.
The proposed framework of large-scale video classification is composed of two phases, training and testing, and it is summarized in Algorithm~ \ref{alg:whole_framwork}.

\begin{algorithm}
	\caption{The large-scale video classification framework.}
	\label{alg:whole_framwork}
	\begin{algorithmic}	
		\REQUIRE $L$ videos for training, corresponds to the labels $L_\mathit{label}$; 
		$T$ testing videos.
		\ENSURE Testing accuracy $T_{acc}$.
		\STATE \textbf{Step 1} TRAINING STAGE: \\
		\quad \textbf{1.1} Extract key frames $\textbf{\text{f}}$ of $\textbf{\text{V}}$ $\leftarrow$ Algorithm~\ref{alg:cdpc}.  \\
		\quad \textbf{1.2} The video class $\text{P}_{\textbf{\text{f}}}$ of each key frame $\leftarrow$ Eq.~\eqref{equ:pos}.\\
		\quad \textbf{1.3} The final classification $\text{R}_\textbf{\text{V}}$ $\leftarrow$ Eq.~\eqref{equ:finalpos}.\\
		\quad \textbf{1.4} Calculate the weights $\text{\textbf{w}}$ for different input $\leftarrow$ Eq.~\eqref{equ:wightsetp1} and \eqref{equ:wightsetp2}.
		
		\STATE \textbf{Step 2} TESTING STAGE:\\
		\quad \textbf{2.1} Extract key frames $\leftarrow$ Algorithm~\ref{alg:cdpc}.  \\
		\quad \textbf{2.2} The $\text{P}_{\textbf{\text{f}}}$ of each key frame $\leftarrow$ Eq.~\eqref{equ:pos}.\\
		\quad \textbf{2.3} The $\text{R}_\textbf{\text{V}}$ $\leftarrow$ Eq.~\eqref{equ:finalpos}.\\
		\quad \textbf{2.4} Computing classification rates of different input.\\
		\quad \textbf{2.5} Combine the weight $\text{\textbf{w}}$ with each rate.\\	
		\RETURN $T_{acc}$.
	\end{algorithmic}
\end{algorithm}

In the training stage, we first obtain the key frames of video $\textbf{\text{V}}$ via Algorithm~\ref{alg:cdpc}, then the LSTM model takes over (step 1.1).
Next, the LSTM model predicts the video class $\text{P}_{\textbf{\text{f}}}$ at each key frame (step 1.2), 
\begin{equation}
\label{equ:pos}
\text{P}_{\textbf{\text{f}}} = \frac{\exp(W_{hc}h_{t,c} + b_c)}{\sum \exp(W_{hc}h_{t,c'} + b_c)},
\end{equation}
and we average these predictions for final classification (step 1.3),
\begin{equation}
\label{equ:finalpos}
\text{R}_\textbf{\text{V}} = \frac{1}{n_c}\sum_{k=1}^{n_c} \text{P}_{\textbf{\text{f}}}.
\end{equation}

Moreover, we propose an approach for the weighted fusion of several inputs networks (i.e., RGB, RGB difference, optical flow and warped flow) that automatically estimates the weight of each input. 
The weights reflect the relevance of each input for the specific video shot~\cite{tang2015gender}.
Note that the proposed weight fusion method is different from the method proposed in~\cite{tang2015gender}.
The method in~\cite{tang2015gender} tries to fusion different patch of the same video, while the proposed in this paper tries to combine different modalities of input features.
First, we obtain $t$ classification rates $\text{\textbf{r}} = \{\text{\textbf{r}}_1, \text{\textbf{r}}_2, ..., \text{\textbf{r}}_t\}$ for different input networks.
We assume that the higher the rate is, the better the feature representation is~\cite{tang2015gender}, then the weights can be calculated as follows:
\begin{equation}
\label{equ:wightsetp1}
t_0 = \frac{\text{\textbf{r}} - \min(\text{\textbf{r}})}{(100 - \min(\text{\textbf{r}}))/10}.
\end{equation}

\begin{table*}
	\centering
	\caption{The key characteristics of the datasets used in the experiments.}
	\begin{tabular}{cccccc} \hline
		Dataset                                  & Resolution       & Classes          & Training                         & Testing                          & Total \\ \hline	
		HMDB51 \cite{Kuehne11}             & $320 \times240$  & 51               & \tht{c}{5,263 \\5,263 \\5,263} & \tht{c}{1,530 \\1,530\\ 1,530} & 6,766  \\ \hline
		UCF101 \cite{soomro2012ucf101}     & $320 \times240$  & 101              & \tht{c}{9,537 \\9,586 \\9,624}  & \tht{c}{3,783 \\3,734\\ 3,696} & 13,320 \\ \hline		
	\end{tabular}
	\label{tab:dataset}
\end{table*}

\begin{table*}
	\caption{Exploration of different input for the proposed framework on the UCF101 dataset.}
		\centering
	\resizebox{1\linewidth}{!}{
		\begin{tabular}{l|cccc|ccc|c} \toprule
			\multirow{2}{*}{Model} & \multicolumn{4}{c}{Input Type}                                               & \multicolumn{3}{|c|}{Weighted} & \multirow{2}{*}{Time (s)}\\ 
			& RGB Image            & RGB Difference & Optic Flow & Warped Flow             & 1/2, 1/2 & 1/3, 2/3 & Ours  & \\ \midrule 		
			Uniform Sampling (8 frames)                   & 54.36     & 52.96          & 58.67      & 56.39       & -        & -        & 55.47 & 1.63\\
			Uniform Sampling (16 frames)                  & 64.39     & 61.59          & 68.36      & 67.25       & -        & -        & 74.29 & 2.32\\
			Uniform Sampling (32 frames)                  & 68.39     & 66.59          & 72.54      & 72.36       & -        & -        & 77.58 & 3.53\\
			K-means~\cite{macqueen1967some}               & 65.35     & 63.58          & 69.69      & 67.96       & -        & -        & 76.67 & 8.54\\
			S-RNN~\cite{sigurdsson2016learning}           & 73.68     & 71.56          & 79.36      & 78.36       & -        & -        & 85.62 & 7.63\\
			Joint Unsupervised Learning~\cite{yang2016joint}& 75.69   & 73.98          & 80.49      & 78.39       & -        & -        & 86.39 & 7.94\\
			Single Frame~\cite{donahue2015long}           & 65.40     & -              & 53.20      & -           & -        & -        & -     & -\\
			Single Frame (ave.)~\cite{donahue2015long}    & 69.00     & -              & 72.20      & -           & 75.71    & 79.04    & -     & -\\
			LRCN-fc$_6$~\cite{donahue2015long}             & 71.12     & -              & 76.95      & -           & 81.97    & 82.92    & -     & -\\
			LRCN-fc$_7$~\cite{donahue2015long}             & 70.68     & -              & 69.36      & -           & -        & -        & -     & -\\ \hline 	
			CNN-TSDPC-LSTM (Ours)   	                          & 79.63     & 78.35          & -          & -           & 80.63    & 82.10    & -     & 4.54\\
			CNN-TSDPC-LSTM (Ours)     	                          & -         & -              & 84.94      & 83.63       & 85.16    & 86.96    & -     & 4.65\\	
			CNN-TSDPC-LSTM  (Ours)    	                          & 79.63     & -              & 84.94      & 83.63       & -        & -        & 93.13 & 4.87\\	
			CNN-TSDPC-LSTM (Ours)    	                          & 79.63     & 78.35          & 84.94      & 83.63       & -        & -        & 91.51 & 4.96\\	\bottomrule
	\end{tabular}}
	\label{tab:input}
\end{table*}

The weight of the lowest rate is set to 1, the other weights are set in the direct proportional to 1 based on the ratios to the lowest rate:
\begin{equation}
\label{equ:wightsetp2}
\begin{aligned}
t_1 &= \text{round}(t_0), \\
t_2 &= \frac{t_0 \times (\max(t_1)-1)}{\max(t_0)} + 1, \\
\text{\textbf{w}}  &= \text{round}(t_2),
\end{aligned}
\end{equation}
where $\text{\textbf{w}}$ is the weight vector of different inputs (step 1.4). 
During the testing stage, key frames of the testing video are selected in the same way as the training stage (step 2.1).
Then we calculate the $\text{P}_{\textbf{\text{f}}}$ and $\text{R}_\textbf{\text{V}}$ in the same way as the training stage as well (step 2.2 and 2.3).
After that, we compute the classification rates of different inputs and combine the rates with the weights $\text{\textbf{w}}$ (step 2.4 and 2.5).
At the end of the framework, the testing accuracy $T_{acc}$ of the whole dataset is returned.

\section{Experiments}

\begin{figure*}
	\centering
		\includegraphics[width=1\linewidth]{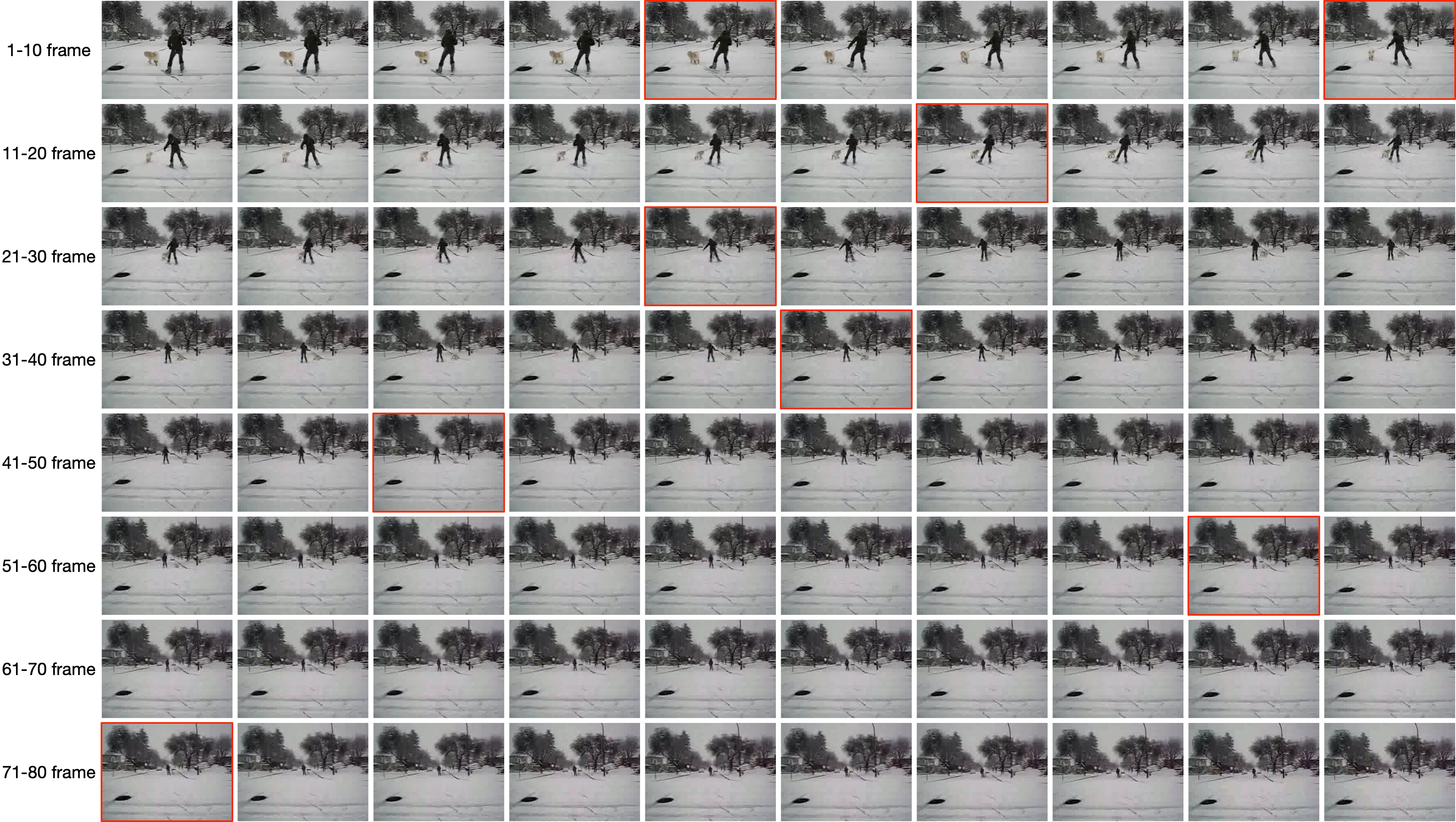}
		\caption{\hao{A video from the UCF101 dataset, which contains 80 frames. The key frames extracted by the proposed method are in red boxes.}}
		\label{fig:keyframe}
\end{figure*}

\begin{table}
	\centering
	\caption{Mean classification performance of the state-of-the-art approaches on the HMDB51 and UCF101 datasets. Methods on the horizontal line are traditional video classification methods and the approaches under the horizontal line  are deep learning methods. \hao{* means the proposed method uses the two-stream I3D as the backbone, and use Kinetics for pre-training.}}
	\resizebox{0.55\linewidth}{!}{
		\begin{tabular}{lrr} \toprule
			Method                                                        & HMDB51 & UCF101 \\ \midrule
			iDT + StackFV \cite{peng2014action}                           & 66.8\% & -      \\ 
			DT + MVSV \cite{cai2014multi}                                 & 55.9\% & 83.5\% \\ 
			iDT + FV \cite{wang2013action}                                & 57.2\% & 85.9\% \\
			iDT + SFV + STP \cite{wang2016robust}                         & 60.1\% & 86.0\% \\
			iDT + HSV \cite{peng2016bag}                                  & 61.1\% & 87.9\% \\
			MoFAP \cite{wang2016mofap}                                    & 61.7\% & 88.3\% \\ 
			iDT + MIFS \cite{lan2015beyond}                               & 65.1\% & 89.1\% \\ \hline
			
			VideoDarwin \cite{fernando2015modeling}                       & 63.7\% & -      \\
			MPR \cite{ni2015motion}                                       & 65.5\% & -      \\
			VGAN \cite{vondrick2016generating}                            &	-      & 52.1\% \\
			Deep Networks, Sports 1M pre-training \cite{karpathy2014large}& -      & 65.2\% \\
			C3D (1 net), Sports 1M pre-training \cite{tran2015learning} & -        & 82.3\% \\
			LRCN (fc$_6$) \cite{donahue2015long}                          & -      & 82.92\%\\
			C3D (3 nets), Sports 1M pre-training \cite{tran2015learning}  & -      & 85.2\% \\
			Res3D \cite{tran2017convnet}                                        & 54.9\%        & 85.8\% \\
			Two Stream \cite{simonyan2014two}                             & 59.4\% & 88.0\% \\ 
			F$_{ST}$CN (SCI fusion) \cite{sun2015human}	                  & 59.1\% & 88.1\% \\	
			Two Stream + LSTM\cite{yue2015beyond}                         & -      & 88.6\% \\ 
			Dynamic Image Networks + IDT \cite{bilen2016dynamic}          & 65.2\% & 89.1\% \\
			AdaScan+Two Stream \cite{kar2016adascan}                      & 54.9\% & 89.4\% \\
			C3D (3 nets) + IDT, Sports 1M pre-training \cite{tran2015learning} & - & 90.1\% \\
			TDD + FV \cite{wang2015action}                                & 63.2\% & 90.3\% \\
			AdaScan+iDT+last fusion \cite{kar2016adascan}                 & 61.0\% & 91.3\% \\		
			TDD + iDT \cite{wang2015action}                               & 65.9\% & 91.5\% \\
			LTC \cite{varol2017long}                                      & 64.8\% & 91.7\% \\
			RGB-I3D, miniKinetics pre-training \cite{carreira2017quo}     & 66.4\% & 91.8\% \\
			Actions Trans \cite{wang2016actions}                          & 62.0\% & 92.0\% \\
			Convolutional Two Stream \cite{feichtenhofer2016convolutional}& 65.4\% & 92.5\% \\ 
			Hybrid-iDT \cite{de2016sympathy}                              & 70.4\% & 92.5\% \\
			KVMF \cite{zhu2016key}                                        & 63.3\% & 93.1\% \\
			AdaScan+iDT+C3D+last fusion \cite{kar2016adascan}             & 66.9\% & 93.2\% \\		 
			TSN (2 modalities, BN-Inception) \cite{wang2016temporal}      & 68.5\% & 94.0\% \\
			Spatiotemporal Multiplier Network \cite{feichtenhofer2017spatiotemporal} & 68.9\% & 94.2\% \\
			TSN (3 modalities, BN-Inception) \cite{wang2016temporal}      & 69.4\% & 94.2\% \\ 
			Cool-TSN \cite{roberto2017procedural}                         & 69.5\% & 94.2\% \\ 
			ST-VLMPF(DF) \cite{dutaspatio}                                & 73.1\% & 94.3\% \\
			Spatiotemporal Pyramid Network \cite{wang2017spatiotemporal}  & 68.9\% & 94.6\% \\
			Spatiotemporal ResNets + IDT \cite{feichtenhofer2016spatiotemporal} & 70.3\% & 94.6\% \\
			Flow-I3D, miniKinetics pre-training \cite{carreira2017quo}    & 72.4\% & 94.7\% \\
			Attention Fusion \cite{long2018multimodal}                         & -         & 94.8\% \\                                         
			Spatiotemporal Multiplier Network + iDT \cite{feichtenhofer2017spatiotemporal} & 72.2\% & 94.9\% \\		 
			RGB-I3D, Kinetics pre-training \cite{carreira2017quo}         & 74.8\% & 95.6\% \\
			Optical Flow guided Feature  \cite{sun2018optical}             & 74.2\% & 96.0\% \\
			Flow-I3D, Kinetics pre-training \cite{carreira2017quo}        & 77.1\% & 96.7\% \\
			Two-Stream I3D, miniKinetics pre-training \cite{carreira2017quo} & 76.3\% & 96.9\% \\
			\hao{I3D RGB + DMC-Net (I3D) \cite{shou2019dmc}} & \hao{77.8\%}  & \hao{96.5\%} \\
			Two-Stream I3D, Kinetics pre-training \cite{carreira2017quo}  & 80.7\% & 98.0\% \\ \hline
			CNN-TSDPC-LSTM (Ours)                                           & 75.52\% & 95.86\% \\  
			\hao{CNN-TSDPC-LSTM* (Ours)} & \textbf{\hao{81.44\%}} & \textbf{\hao{98.45\%}} \\ \bottomrule
		\end{tabular}}
		\label{tab:state}
	\end{table}

\begin{table}
	\centering
	\caption{Compression Ratio (\%) on HMDB51 and UCF101 datasets.}
	\begin{tabular}{lcccccr} \toprule
		Dataset & \#Total Frame & \#Avg. Frame & \#Total Key Frame & CR(\%) \\ \midrule	
		HMDB51  & 634,552              & 93.8                 & 108,256                  & 82.94 \\ 
		UCF101  & 2,485,519            & 186.6                & 213,120                  & 91.43 \\  \bottomrule		
	\end{tabular}
	\label{tab:compression_ratio}
\end{table}

To evaluate the effectiveness of the proposed method, we conduct experiments with two popular public datasets. More comparisons are shown in Table~\ref{tab:dataset}. 

\subsection{Datasets}
\noindent \textbf{HMDB51} dataset\footnote{\url{http://serre-lab.clps.brown.edu/resource/hmdb-a-large-human-motion-database/}} 
contains a total of 6,766 video clips distributed in a large set of 51 action categories collected from various sources, mostly from movies, public datasets and YouTube. 
Each category contains a minimum of 101 video clips. 
The videos are taken with the resolution of $320 \times 240$ with 30 fps.

\noindent \textbf{UCF101} dataset\textsc{}\footnote{\url{http://crcv.ucf.edu/data/UCF101.php}} is a widely-used dataset for action recognition. 
It comprises of realistic videos collected from YouTube lasting 7 sec. on average. 
Videos have a spatial resolution of $320 \times 240$ pixels with 25 fps.
UCF101 gives the largest diversity in terms of actions and with the presence of large variations in object appearance, scale and pose, camera motion, viewpoint, cluttered background, illumination conditions.

\subsection{Setups}
For a fair comparison, we set $t = 0.2$ as in~\cite{rodriguez2014clustering}.
All the experiments are run at Ubuntu with 2 TITAN Xp GPUs.
For one video, the selection process is quite fast with the help of GPUs. \hao{We follow existing methods (e.g., \cite{carreira2017quo}) and train our model by using cross-entropy loss.}

\subsection{Experimental Results}
We present extensive experimental results to demonstrate the necessity and efficiency of the proposed method and framework on large-scale video classification task.

\noindent \textbf{Comparison of other Extraction Methods.}
We first compare different key frames extraction methods on large-scale video classification task.
The column ``Input Type'' of Table~\ref{tab:input} shows the performance comparison between our method and the uniform sampling method with different rate (8, 16, 32 frames), k-means~\cite{macqueen1967some}, S-RNN~\cite{sigurdsson2016learning}, Joint Unsupervised Learning~\cite{yang2016joint}, single frame~\cite{donahue2015long} and LRCN~\cite{donahue2015long} on the UCF101 dataset (split 1).
Note that we follow \cite{wang2016temporal} and adopt four different information, i.e., RGB image, RGB difference, optic flow and warped flow as inputs.
The results show that our method CNN-TSDPC-LSTM consistently outperforms all the baselines with signification improvements, which validates that there is significant informative motion and spatial information available around key frames.
\hao{Finally, to better understand and evaluate the proposed key frame extraction method, we show one video example from UCF101 in Figure \ref{fig:keyframe}.}

\noindent \textbf{Combination with Weight Strategy.}
We then add the proposed weight strategy to our framework to test whether they benefit for this task. 
As shown in the column ``Weighted'' of Table \ref{tab:input}, the proposed weight method outperforms the baseline method LRCN~\cite{donahue2015long} when two inputs (RGB Image and Optic Flow) are employed.
Note that ``1/2, 1/2'' and ``1/3, 2/3'' are proposed in \cite{donahue2015long} and are two different weight settings which represent the fixed weights of two different inputs.
Besides, when we adopt three inputs, i.e., RGB image, optic flow and warped flow, we achieve the best performance compared with other baselines. 
However, we observe that introduce RGB difference will degrade the performance, which is also observed in TSN~\cite{wang2016temporal}.

\noindent \textbf{Comparison with State-of-the-Art Methods.}
We assemble these three inputs and all the techniques described as our final video classification method, and test it on both HMDB51 and UCF101 datasets.
For both HMDB51 and UCF101 datasets, we compare the proposed framework i.e., CNN-TSDPC-LSTM-Fusion, with the state-of-the-art traditional approaches, e.g., iDT + HSV~\cite{peng2016bag} and MoFAP~\cite{wang2016mofap}.
We also compare the proposed method with deep learning representation methods such as LRCN \cite{donahue2015long}, KVMF~\cite{zhu2016key} and TSN~\cite{wang2016temporal}.
The results are summarized in Table~\ref{tab:state}, our method achieves competitive classification accuracy compared with these methods. 
Note that the classification performance of the proposed framework is worse than several of baselines such as Two-Stream I3D \cite{carreira2017quo}.
Carreira and Zisserman \cite{carreira2017quo} achieve the best performance after pre-training on extra data, i.e., Kinetics.
However, the proposed method is more efficient and has a higher compression ratio which is defined as the relative amount of savings provided by the summary representation.
The definition of the compression ratio is, 
$\text{CR}(\textbf{\text{V}}) = 1 - \frac{n_c}{N},$
where $n_c$ and $N$ are the number of key frames and the number of frames in the original video $\textbf{\text{V}}$ respectively. 
The results are shown in Table~\ref{tab:compression_ratio}.
Generally, a high compression ratio means for a compact video summary and also means less video processing time.
\hao{Finally, to further prove the effectiveness of our proposed method, we use the two-stream I3D as our backbone, and use Kinetics for pre-training. As can be seen from Table \ref{tab:state}, this model achieves state-of-the-art results on both datasets.}

\begin{table}
	\centering
	\caption{Time comparison of different models on UCF101 dataset.}
	\begin{tabular}{lc} \toprule
		Model & Time \\ \midrule		
		CNN-TSDPC-LSTM (Ours)   & 1.00  \\ 
		Optical Flow guided Feature~\cite{sun2018optical} & 1.13x \\ 
		Two-Stream I3D~\cite{carreira2017quo}  & 1.35x   \\ 	\bottomrule	
	\end{tabular}
	\label{tab:time}
\end{table}

\begin{table}
	\centering
	\caption{\hao{Computational complexity of different networks}.}
	\begin{tabular}{cc} \toprule
		\hao{Network Architecture} & \hao{GFLOPs} \\ \midrule
		\hao{C3D \cite{tran2015learning}} & \hao{38.5} \\
		\hao{Res3D-18 \cite{tran2017convnet}} & \hao{19.3} \\
		\hao{ResNet-152 \cite{he2016deep}} & \hao{11.3} \\
		\hao{ResNet-18 \cite{he2016deep}} & \hao{1.78} \\
		\hao{PWC-Net \cite{sun2018pwc}} & \hao{36.15} \\
		\hao{CNN-TSDPC-LSTM (Ours)} & \hao{1.45}\\ \bottomrule
	\end{tabular}
	\label{tab:flpos}
\end{table}

\noindent \textbf{Efficiency Comparison.}
We also measure the average classification time on UCF101 dataset. 
The column ``Time'' of Table \ref{tab:input} lists the average time of different sampling approaches. 
The results showing that we can obtain high classification accuracy while keeping the complexity low compared with the state-of-the-art key frame extraction methods such as S-RNN~\cite{sigurdsson2016learning} and Joint Unsupervised Learning~\cite{yang2016joint}.
Consequently, the proposed method is time efficient both on-line and off-line, and can be readily adopted in real-world applications.
Moreover, as we expected we observe that more inputs consumer more times.
We also provide time comparison with the state of the art methods on UCF101 dataset.
Results are shown in Table \ref{tab:time}, we can see that the proposed method is faster than Optical Flow guided Feature~\cite{sun2018optical} and Two-Stream I3D~\cite{carreira2017quo}.
\hao{In Table \ref{tab:flpos}, we also provide the GFLOPs results compared with other network architectures for video analysis, including ResNet-18 \cite{he2016deep}, ResNet-152 \cite{he2016deep}, C3D \cite{tran2015learning}, and Res3D \cite{tran2017convnet}.
We observe that the complexity of the proposed method is smaller compared to that of other architectures, which
makes it run much faster.}

\section{Conclusion}

We present an unsupervised key frame extraction method, i.e., TSDPC, for video key frame extraction, which can greatly reduce the redundant information of the original video while preserve the temporal information through selecting the number of key frames automatically.
In addition, a framework CNN-TSDPC-LSTM aimed at large-scale video classification which consists of CNN, TSDPC and LSTM is proposed. 
Moreover, a fusion strategy of different input networks is presented to boost the accuracy of video classification.
Experimental results on a variety of public datasets demonstrate that (i) our framework is capable of summarizing video efficiently regardless of the visual content of videos.
(ii) our framework is with high efficiency while the mean classification accuracy is comparable with other state-of-the-art methods.
Note that the proposed method is generic, which thus can be beneficial to other video processing tasks.

\begin{acks}
This work is supported by the PRIN project PREVUE (Prot. 2017N2RK7K) and by the EU H2020 project AI4Media under Grant 951911.
\end{acks}

\bibliographystyle{ACM-Reference-Format}
\bibliography{sample-base}

%
%
%
%
%
%
%
%

\end{document}